\documentclass[letterpaper, 10 pt, conference]{ieeeconf}  

\IEEEoverridecommandlockouts                              

\usepackage{graphics} 
\usepackage{epsfig} 
\usepackage{times} 
\usepackage{amsmath} 
\usepackage{amssymb}  
\usepackage{amsfonts}       
\usepackage{amstext} 
\usepackage{mathtools}
\usepackage{bbm}
\usepackage{float}

\usepackage{hyperref}       

\usepackage{cite}
\usepackage[capitalize, nameinlink]{cleveref} 
\usepackage{booktabs}       
\usepackage{makecell}
\usepackage{siunitx}
\usepackage{algorithm}
\usepackage[noend]{algorithmic}
\usepackage{threeparttable}

\newcommand{\CASE}[1]{\STATE \textbf{case} #1\textbf{:} \begin{ALC@g}}
\newcommand{\ENDCASE}{\end{ALC@g}}

\newcommand{\DEFAULT}{\STATE \textbf{default:} \begin{ALC@g}}
\newcommand{\ENDDEFAULT}{\end{ALC@g}}
\newcommand{\DEFAULTLINE}[1]{\STATE \textbf{default:} }
\usepackage[nolist]{acronym}
\usepackage[dvipsnames]{xcolor}

\usepackage{hyphenat} 
\usepackage{xspace}

\title{\LARGE \bf
Downwash-aware Configuration Optimization for Modular Aerial Systems
}
\author{Mengguang Li and Heinz Koeppl
\thanks{*This work has been funded by the LOEWE initiative (Hesse, Germany) within the emergenCITY center [LOEWE/1/12/519/03/05.001(0016)/72].}
\thanks{The authors are with the Department of Electrical Engineering and Information Technology, Technische Universität Darmstadt, 64287 Darmstadt, Germany.
{\tt\small \{mengguang.li, heinz.koeppl\}@tu-darmstadt.de}}%
}
\begin{document}
\maketitle
\thispagestyle{empty}
\pagestyle{empty}
\begin{abstract}
This work proposes a framework that generates and optimally selects task-specific assembly configurations for a large group of homogeneous modular aerial systems, explicitly enforcing bounds on inter-module downwash. Prior work largely focuses on planar layouts and often ignores aerodynamic interference. In contrast, firstly we enumerate non-isomorphic connection topologies at scale; secondly, we solve a nonlinear program to check feasibility and select the configuration that minimizes control input subject to actuation limits and downwash constraints. We evaluate the framework in physics-based simulation and demonstrate it in real-world experiments.
\end{abstract}
\section{INTRODUCTION} \label{sec1}
Recent advances in modular reconfigurable robotic systems show promising results, due to their high design flexibility compared with conventional systems \cite{annurev023834}. By reassembling and reusing a group of repeated modules, these systems enable robust, resilient and cost-efficient solutions, thereby offering versatility and scalability for task-specific adaptation. 

In aerial robotics, many modular systems have emerged over the last decade. In this work, we focus on rotor-based actuation. Early work on modular multirotors demonstrated assemblies built from single-rotor modules \cite{5509882} and mid-air self-assembly of standard quadrotors \cite{8461014}. These elegant designs have limitations though, as the assemblies are largely planar, rotor orientations remain collinear, and therefore the overall system is underactuated, which restricts the application scope. In general, a common strategy is to tilt rotors to break collinearity. Rotors can be either fixed-tilt \cite{0278364919856694} or actively tilting during flight\cite{0278364920943654}.

When tilting the rotors, the aerodynamic effect of downwash\cite{5569026}, which is the deflection of air under the propellers, needs to be considered, as neighboring wakes may interfere. Airflow interference between rotors can reduce efficiency and even induce system instability, effects that are pronounced especially in modular designs where rotors are often in close proximity. Downwash has, for example, already been considered in the design of aerial platforms\cite{7139851}. Researchers have also proposed control allocation strategies to reduce the downwash effect in overactuated aerial systems\cite{10214628}. A recent study shows that the quadrotor-induced combined airflow can be well-approximated as a turbulent jet\cite{10804051}.

\begin{figure}[t]
    \centering
    \includegraphics[width=.8\linewidth]{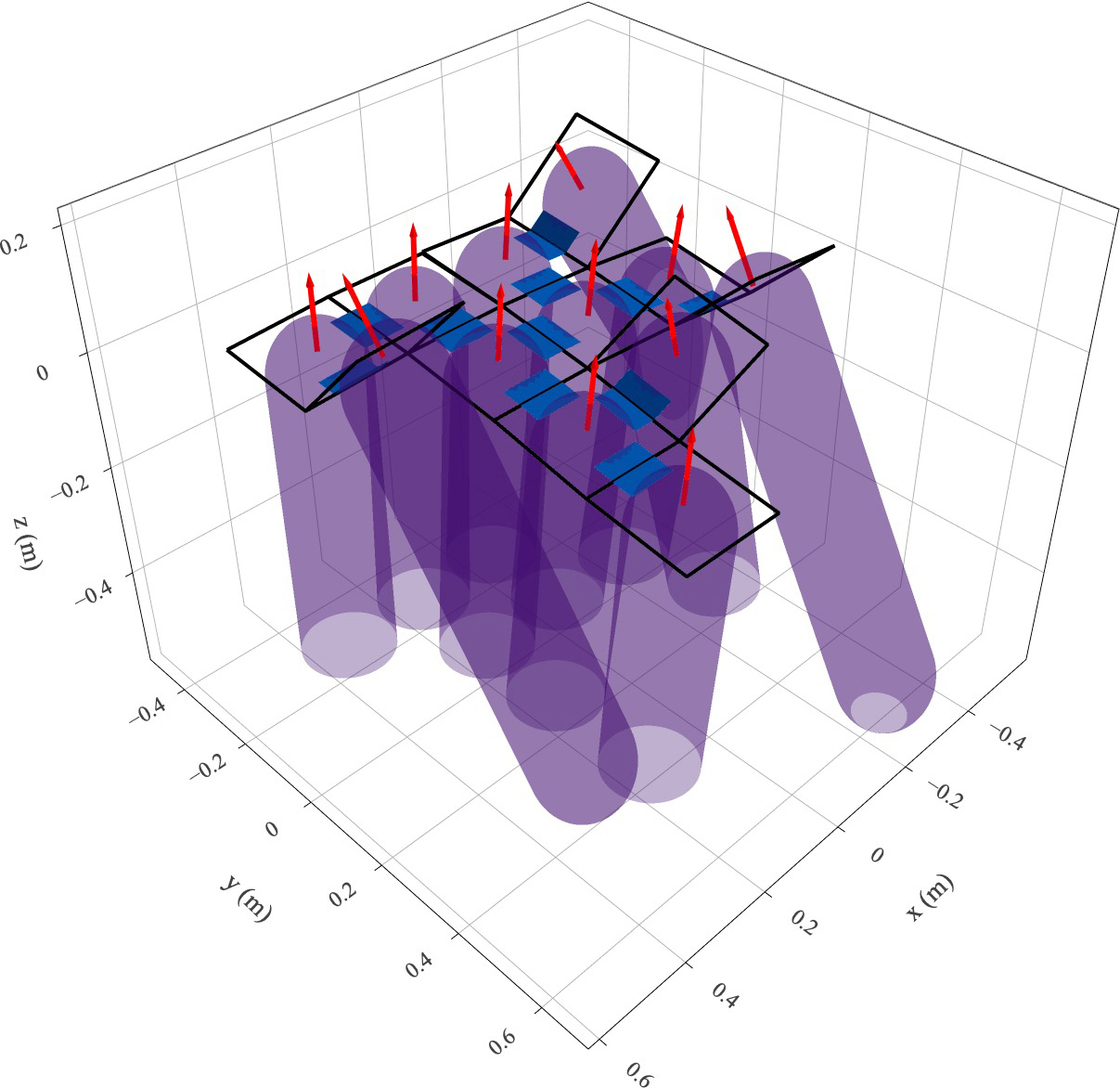}
    \caption{The optimal assembly of $12$ modules, given the target wrench set defined in \cref{sec:4}. Each black square represents one module, each blue rectangle represents a rigid connection, each red arrow represents the module frame $z$-axis, and the purple capsules indicate the collision-free downwash volume for each module.}
    \label{fig:fig1}
\end{figure}
However, in the design of modular aerial systems, the downwash effect is rarely explored or sometimes ignored completely. Interference can occur within a single module or between modules. For instance, in \cite{10160555}, each module is a multirotor with four fixed-tilt rotors, where inter-module downwash interference is ignored, potentially affecting system performance. In \cite{9636086}, a standard quadrotor is attached via a passive hinge joint to a cage to form a module, thereby extending the overall system actuation capability. This additional one degree of freedom (DOF) may cause downwash interference within the configuration but was not considered. Likewise, in \cite{10801469}, each quadrotor-based module is extended with a $2$-DOF passive gimbal, further enlarging the actuation space, again without considering downwash effects.

Therefore, this paper proposes a design and optimization framework that addresses these gaps. We first use the standard quadrotor as the building block of our base module so that within each module, the rotor downwash interaction can be neglected. Inspired by \cite{10610414}, we allow non-planar assemblies to explore fully 3D arrangements while respecting each module's airflow envelope. Given a task-specific wrench requirement, our method automatically determines the number of modules needed, the inter-module connection topology, and their 3D spatial arrangement to satisfy the requirement with minimum total effort. The pipeline explicitly accounts for rotor downwash interference by enforcing collision-free spatial volume, yielding physically realizable layouts. An example is shown in \cref{fig:fig1}.

This paper is organized as follows. \cref{sec:2} defines the applicable module within this framework and states the problem. \cref{sec:3} presents the enumeration algorithm for non-isomorphic configurations. \cref{sec:4} details configuration optimization and \cref{sec:5} describes the control. Physics-based simulation and real-world experiments are reported in \cref{sec:6}, and \cref{sec:7} concludes this work.
\section{MODULES AND PROBLEM STATEMENT}\label{sec:2}
Specifically, we use an X-layout quadrotor with collinear, unidirectional propellers as the module’s building block. This platform is widely used both commercially and in research, including agile vision-based flight \cite{abl6259}, swarm exploration \cite{10161498}, and even aerial physical interaction despite its underactuated nature \cite{6907782}. Consequently, commercially available components can be readily adapted to our method.
\subsection{Module and assembly}
\begin{figure}[t]
    \centering
    \includegraphics[width=1.0\linewidth]{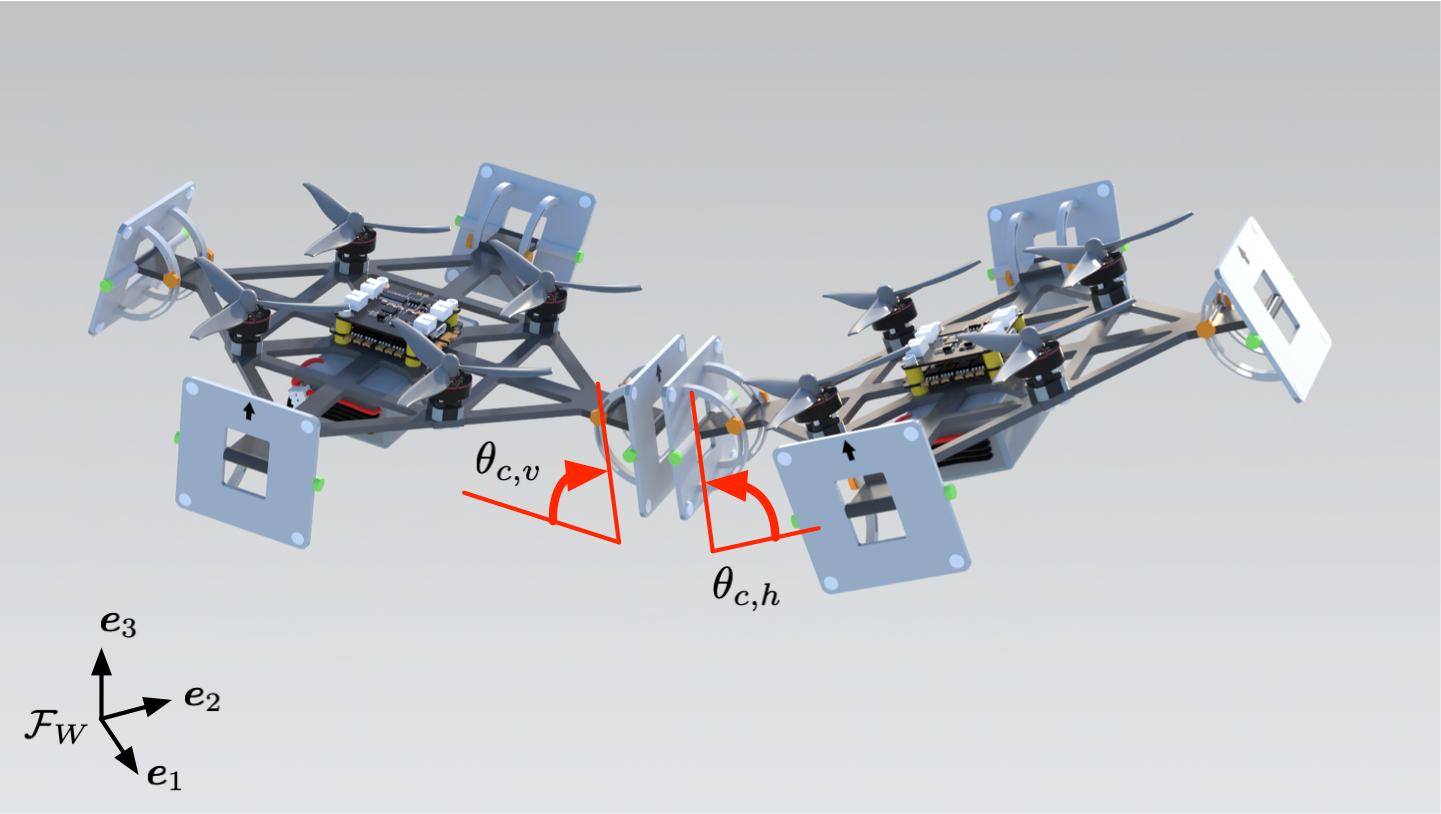}
    \caption{An illustration of two modules with four connection ports on each frame. Each connector can rotate along the green axis, where the orange screw can fix the connector. The white circles on the connector are permanent magnets.}
    \label{fig:design}
\end{figure}

Each module has the same arm length and is equipped with four identical connection ports as shown in \cref{fig:design}. Two modules can rigidly connect to each other using one of their connectors by aligning the contact areas, as indicated by the arrow on the connector. For each connector on the module $v$, the angle $\theta_{c,v}$ of connector $c_v\in\{1,2,3,4\}$ can be adjusted within the prescribed range $[\pi/4,3\pi/4]$. One realization of the connector is shown in \cref{fig:design}, where the four white circles represent permanent magnets, as in standard modular systems \cite{8461014}. Other position-adjustable hinge joints can also be used, as in \cite{10610414} with a 3D-printed connector for each fixed angle. Note that when all angles $ \theta_{c,v} $ are fixed at $\pi/2$, the system resembles that in \cite{8461014}. The advantage of this type of connection is the significant enlargement of configuration space compared with purely planar docking, by modifying the connector angle. By adjusting the connection angle, the system can break the collinearity of the rotor orientation, resembling the standard fixed-tilt rotor design and, at the same time, the downwash effect within a single module is not affected; only inter-module effects need to be considered. With this, the structure is able to reconfigure into non-planar layouts and adapt to various tasks. Once the task-specific configuration is given, the connector angles are fixed for that structure, and the whole system can be considered as a rigid body, which does not suffer from the common drawbacks of actuated tiltrotor design, such as additional actuation complexity, servo-motor induced rotational limitations, and singularities\cite{0278364920943654}. 
We consider a modular aerial system composed of $n$ identical base modules. We assume that assemblies are acyclic, which have no kinematic loops, and that there is at most one physical connection between any pair of modules. Excluding cycles avoids coupled connector angles that overconstrain the structure.

\subsection{Problem statement} \label{sec:problem}
We denote a configuration as a tuple $\mathcal{C} \;:=\; \bigl(n,\; \mathcal{G},\; \Lambda),$ where $\mathcal{G}=(V,E)$ is a finite, edge-labeled undirected tree with $|V|=n$, each vertex \(v\in V\) is a module instance, and each edge \(e = \{ v,h \}\in E\) represents exactly one physical connection. The edge label is denoted as $\Lambda(e)=(c_v,c_h,\theta_e)$, where $c_v, c_h$ are the connectors of the two modules connected by edge $e$, and $\theta_e = (\theta_{c,v} + \theta_{c,h}) \in [ \pi/2, 3\pi/2 ] $ is the relative angle between the two module frames. We set the angle $\theta_{c,v}=\theta_{c,h}$ for each connection.

We model the task requirements as a finite set of $K$ target wrenches
\begin{equation*}
    \mathcal{W} = \{ \boldsymbol{b}_k  \in \mathbb{R}^6 | k \in \{1,\dots,K\}\},
\end{equation*}
each entry $\boldsymbol{b}_k$ is a wrench vector, composed of a force vector $\boldsymbol{b}_{k,f} \in \mathbb{R}^3$ and a torque vector $\boldsymbol{b}_{k,\tau} \in \mathbb{R}^3$, that the system must generate within its actuation limits. Our objective is to find, for a given set of target wrenches $\mathcal{W}$, a configuration $\mathcal{C}$ that satisfies the wrench set while minimizing the overall control input. 

The problem stated above is not trivial to solve. The number of tree topologies grows exponentially in $n$, and connector port choices further increase the search space by an additional exponential factor. Furthermore, each edge on each topology carries a continuous angle $\theta_e$, yielding a mixed discrete–continuous design space that further complicates direct solution.

To manage this complexity, we use a two-stage pipeline. Firstly, we set all connector angles temporarily fixed at $\pi/2$, enumerate all non-isomorphic configurations for $n$ modules. Secondly, given a task wrench set $\mathcal{W}$, we evaluate each configuration by searching the connector angle space to check feasibility and compute cost of all the control input. We then select the configuration with the smallest cost.
\section{Enumeration of Non-Isomorphic Configurations}\label{sec:3}
In this section, we solve the first part stated above. Setting all the connection angles to $\pi/2$ reduces the task to the classical problem of square-lattice modular robots enumeration, a problem that has been widely studied. We extend the matrix-based approach in \cite{doi:10.5772/10489} to handle identically shaped assemblies that may differ in their topological interconnections, as in our case. To address scalability, we also introduce a sampling-based heuristic that trades completeness for tractability when the number of modules becomes large.

\subsection{Exhaustive enumeration for small \texorpdfstring{$n$}{n}} \label{sec:exhaustive}
\begin{figure}
    \centering
    \includegraphics[width=0.65\linewidth]{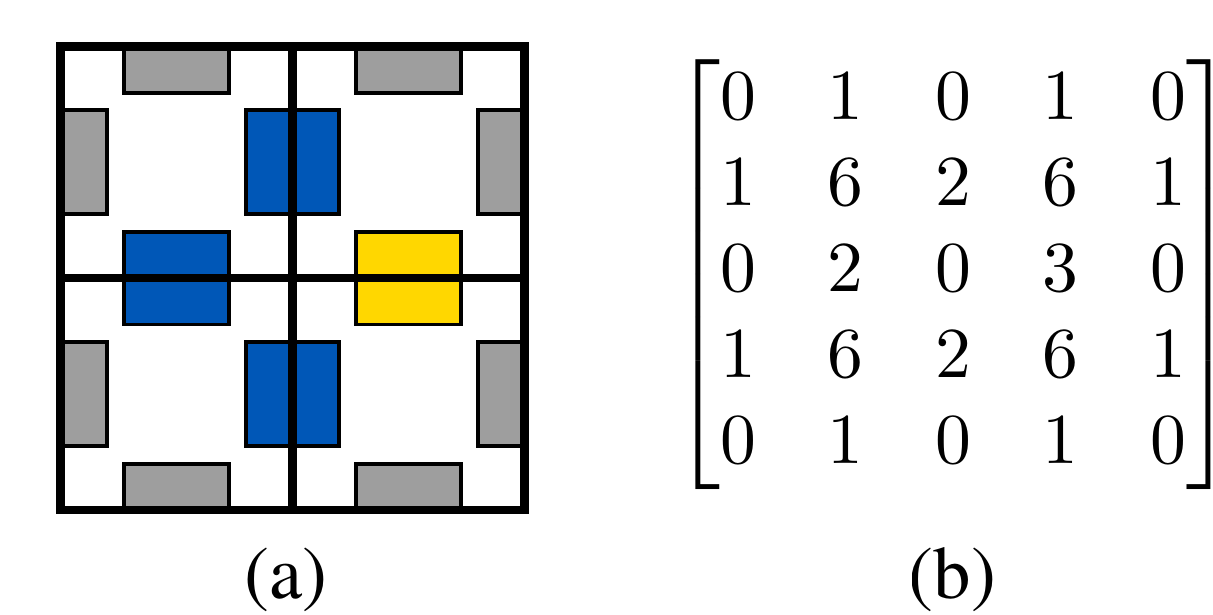}
    \caption{An example of a four-module assembly and its matrix representation, where a black square represents a single module, the four rectangles within each square are the connectors. Blue indicates a valid connection, yellow indicates a blocked connection, and gray indicates an available connector.}
    \label{fig:single-module}
\end{figure}

\begin{figure*}
    \centering
    \includegraphics[width=1.0\linewidth]{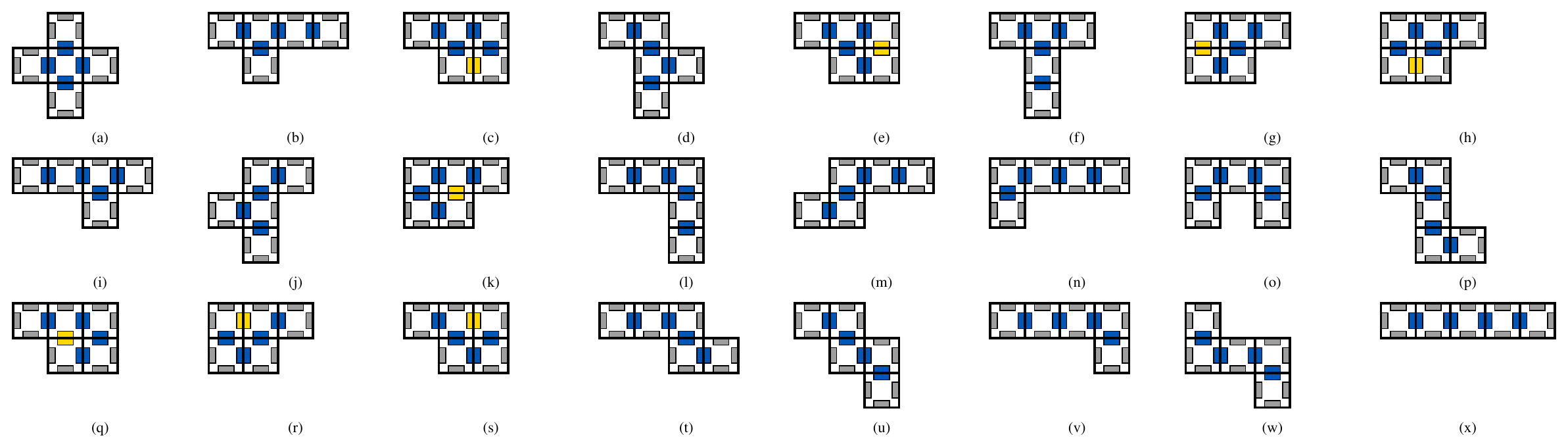}
    \caption{All non-isomorphic configurations under the symmetric point group $C_4$ using $5$ modules.}
    \label{fig:non-iso}
\end{figure*}

We represent each module as a $ 3 \times 3 $ matrix as in \cite{doi:10.5772/10489}, where the module center is coded as \texttt{6}, and each available connector on the cardinal edges is coded as \texttt{1}. Two modules may be joined when opposing edge entries (both \texttt{1}) are aligned; a valid connection is recorded by summing the interface entries and coded as \texttt{2}. As only one connection between two modules is allowed, introducing one new module may block other available connectors. Any edge rendered unusable by the connection is recoded as \texttt{3}. An illustration of the matrix representation is given in \cref{fig:single-module} where four modules are connected together.
For enumeration, we start with a single module and add a new module to all available connectors. For each newly generated configuration, we identify whether it is already known or isomorphic to known configurations. For our system, we consider configurations under the symmetric point group $C_4$ to be isomorphic, i.e., the four rotations around the module's local frame $z$-axis, $\{ 0^{\circ}, 90^{\circ},180^{\circ},270^{\circ} \}$. As each module produces directional thrust along its positive body frame $z$-axis, flip symmetry does not apply here. With this, all non-isomorphic configurations can be found by visiting all available connectors. Unlike \cite{doi:10.5772/10489}, with blocked connections the matrix representation is able to distinguish structures with the same shape but different connection topologies, as changes in the connection angles can result in different configurations for the same assembly shape. An illustration of all non-isomorphic configurations with $5$ modules is shown in \cref{fig:non-iso}. 
A table of all non-isomorphic configurations up to $11$ modules is given in \cref{tab:exhaustive-enum}. Experiments are conducted on Gentoo Linux (kernel \texttt{6.12.21-gentoo}) with an AMD EPYC 7702P processor. Runtime is measured single-threaded using one logical CPU, $2.00$ GHz. The algorithm can be generalized to other module families by adapting the rule of blocked connections and the symmetry group used for isomorphism check. By defining a configuration signature to represent a set of isomorphic configurations\cite{6696972}, the runtime can be improved, but the exponential scaling remains the same and the exhaustive procedure for large $n$ becomes computationally impractical.
\begin{table}[t]
\centering
\caption{Exhaustive enumeration of non-isomorphic configurations for connector angles all fixed to $\pi/2$.}
\label{tab:exhaustive-enum}
\begin{threeparttable}
\setlength{\tabcolsep}{6pt}
\begin{tabular}{@{}
    S[table-format=2.0]  
    S[table-format=6.0]  
    S[table-format=6.0]  
    S[table-format=2.4] 
@{}}
\toprule
\multicolumn{1}{c}{$n$} &
\multicolumn{1}{c}{\makecell{Number of\\available\\configurations}} &
\multicolumn{1}{c}{\makecell{Number of\\non-isomorphic\\configurations}} &
\multicolumn{1}{c}{\makecell{CPU\\Time (s)}} \\
\midrule
1 &     1 &     1 &   0.0000 \\
2 &     4 &     1 &   0.0006 \\
3 &     6 &     2 &   0.0007 \\
4 &    16 &     7 &   0.0017 \\
5 &    68 &    24 &   0.0072 \\
6 &   272 &    97 &   0.0261 \\
7 &  1242 &   401 &   0.1147 \\
8 &  5740 &  1772 &   0.5488 \\
9 & 27960 &  7930 &   2.7293 \\
10 & 136628 & 36335&  14.1008 \\
11 & 678204 & 168249& 78.5251\\
\bottomrule
\end{tabular}
\end{threeparttable}
\end{table}

\subsection{Heuristic sampling for large \texorpdfstring{$n$}{n}}
For large $n$, we explore the physical characteristics of the system and propose a sampling-based enumeration that trades off completeness for computational feasibility.

In general, for the same number of modules $n$, all the configurations have the same total weight, but the radius of gyration can vary depending on how widely the modules are distributed. A smaller radius of gyration indicates a lower moment of inertia, which leads to a lower control input for rotational dynamics, resulting in a compact structure with higher maneuverability. For our aerial system, we therefore encourage compact configurations, where the mass is distributed not too far from the center of mass (COM). As has been shown, a chain-like configuration reaches actuation saturation earlier \cite{9196735}. The matrix representation already encodes the spatial information of the structure. We consider all entries equal to \texttt{6} in the matrix, which represent the center of each module. It is then straightforward to compute the COM for a given configuration as $\bar{\boldsymbol{g}}_m \in \mathbb{R}^2_{\geq 0}$, which is the average coordinates of all the \texttt{6} in the matrix. To scale up, we sample configurations with lower radius of gyration with higher probability. For $n$ modules, we sample $N_{n} \in \mathbb{N}_{>0}$ from the $M_n$ available non-isomorphic configurations $\mathcal{S}_n=\{\mathcal{C}_1(n), \dots, \mathcal{C}_{M_n}(n)\}$. If $M_n\leq N_n$, we keep all configurations without sampling. For $M_n > N_n $, we calculate the radius of gyration with
\begin{equation*}
    g_m\;=\;\sqrt{\frac{1}{n}\sum_{i=1}^{n}\big\lVert \boldsymbol{g}_{m,i}-\bar{\boldsymbol{g}}_m\big\rVert^2},
\qquad 
\bar{\boldsymbol{g}}_m=\frac{1}{n}\sum_{i=1}^{n}\boldsymbol{g}_{m,i},
\end{equation*}
where $\{\boldsymbol{g}_{m,i}\}_{i=1}^n$ are the module-center coordinates of the $m$-th configuration $\mathcal{C}_m(n)$, $m \in \{1, \dots,M_n \}$. Let $g_{m,0}:=\min_{1\le m\le M_n} g_m,$ we set for each $m$-th configuration a Gaussian weight $w_m$ and its sampling probability $P_m$ as 
\begin{equation*}
    w_m:=\exp\!\big(-\frac{(g_m-g_{m,0})^2}{2\sigma_n^2}\big), \quad P_m :=\frac{w_m}{\sum_{m'=1}^{M_n}w_{m'}}.
\end{equation*}
We can now draw $N_n$ configurations without replacement from the set $\mathcal{S}_n$ using the probability $P_m$. After sampling, we proceed with the same steps as above in the exhaustive search. The algorithm is summarized in \cref{algo:sampling}. 
\begin{algorithm}[t]
\caption{Sampling-based enumeration of non-isomorphic configurations}
\label{algo:sampling}
\begin{algorithmic}[1]
\STATE \textbf{Init:} $n$, $\{N_i\}_{i=2}^{n}$, $\{\sigma_i\}_{i=2}^{n}$, $\mathcal{S}_1 \gets \{\mathcal{C}_1(1)\}$,\; $M_1 \gets 1$
\FOR{$i=2$ \TO $n$}
    \FOR{$m=1$ \TO $M_{i-1}$}
        \FOR{connector $c$ coded as \texttt{1} in $\mathcal{C}_m(i-1)$}
            \STATE $\mathcal{C}'(i) \gets$ attach one new module at $c$
            \IF{$\mathcal{C}'(i)$ (up to symmetry group $C_4$) $\notin \mathcal{S}_i$}   
            \STATE $\mathcal{S}_i \gets \mathcal{S}_i \cup \{\mathcal{C}'(i)\} $
            \ENDIF
        \ENDFOR
    \ENDFOR
    \IF{$M_i \le N_i$}
        \STATE \textbf{return} $\mathcal{S}_i$
    \ELSE
        \STATE compute $\bar{\boldsymbol{g}}_m, g_m,g_{m,0},w_m,P_m$
        \STATE $\mathcal{S}'_i \gets \textsc{Sample}(\mathcal{S}_i, N_i; P_m)$ \textit{(without replacement)}

        \STATE \textbf{return} $\mathcal{S}_i \gets \mathcal{S}'_i$
    \ENDIF
\ENDFOR
\STATE \textbf{return} $\mathcal{S}_2, \dots,\mathcal{S}_n $
\end{algorithmic}
\end{algorithm}

\begin{figure}[t]
    \centering
    \includegraphics[width=1.0\linewidth]{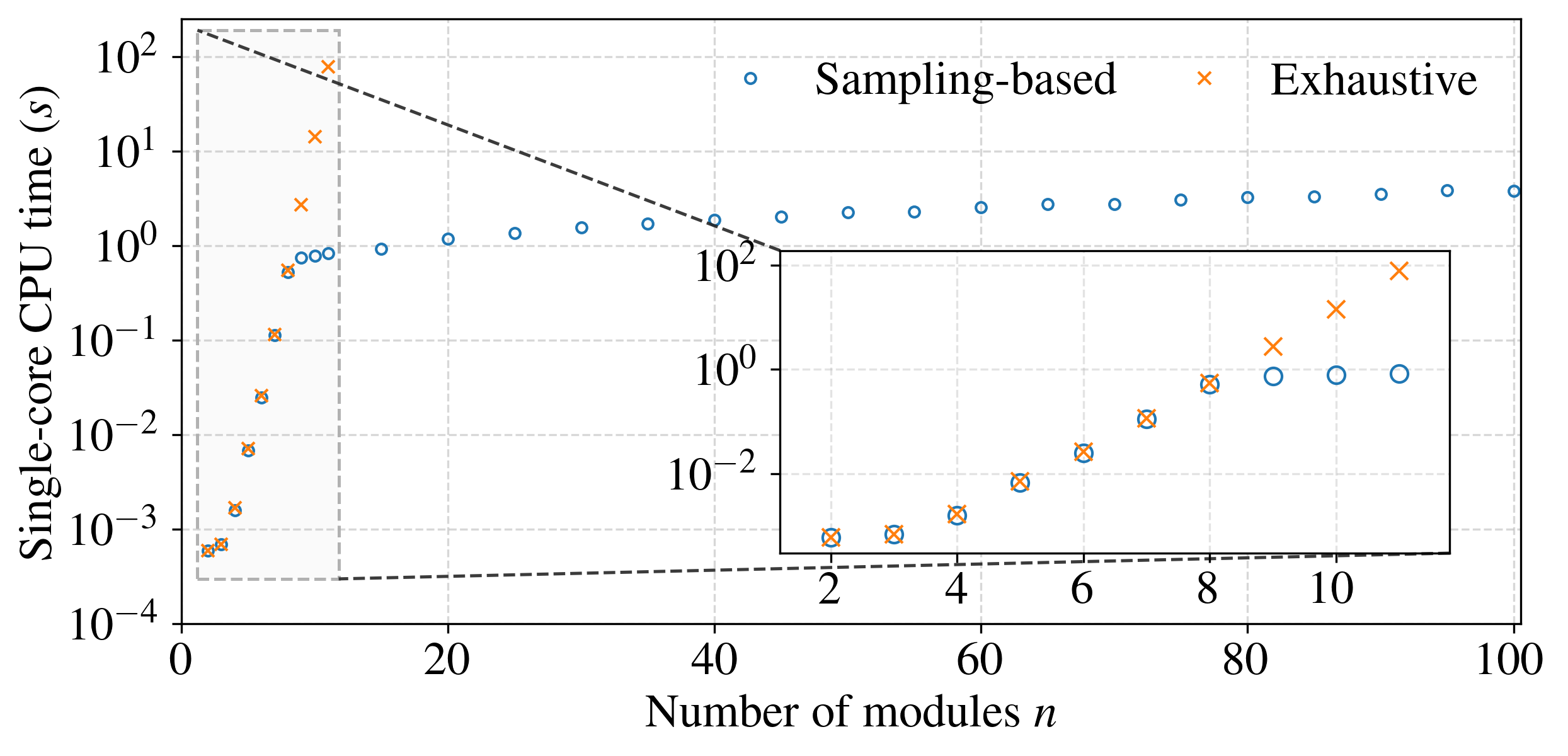}
    \caption{Comparison of exhaustive and sampling-based enumeration. The sample count is fixed at $N_n = 500$ for all $n$. For $n > 11$, CPU time is shown only for every fifth value of $n$.}
    \label{fig:enum_sample}
\end{figure}

With this, configurations with a large number of modules can be explored within a manageable runtime. The computation time per number of modules is given in \cref{fig:enum_sample}, and the hardware is the same as in \cref{sec:exhaustive} with one CPU-core. As can be seen, the method is scalable, as the run time grows approximately linearly with a growing number of modules. The trade-off is completeness, for larger $n$, the fraction of sampled assemblies relative to the rapidly expanding set of all non-isomorphic configurations drops approximately exponentially. By changing the variance $\sigma_n$ of the Gaussian weights, the aggressiveness of sampling can be adjusted. To retain useful coverage, we bias configurations by the radius of gyration of the COM, where smaller $g_m$ receive higher probability based on the Gaussian kernel. This prioritizes compact assemblies while still exploring diverse topologies.

For other robotic systems, different sampling schemes can be tailored to robot-specific properties. For example, uniform sampling explores growth without added bias, or reversed sampling as ours by defining $ g_{m,0}:=\max_{1\le i\le M_n} g_m $, which encourages chain-like or more spatially spread out configurations.
\section{Configuration Optimization} \label{sec:4}
From the previous section, we obtain, for each $n$, a set of non-isomorphic candidate configurations, where all connector angles are set to $\pi/2$. We now propose an optimization procedure that, given a set of task wrenches, selects connector angles and input allocations to meet the wrench requirements while accounting for downwash interactions. Concretely, we impose geometric constraints that prevent harmful overlap of the rotor downwash regions between modules, and we solve for feasible, low control input configurations that satisfy the tasks. Afterward, we compare across the configurations to select the best variant among them.

\subsection{Configuration kinematics}
We first derive the kinematics for any non-isomorphic configuration of $n$ modules, using the matrix representation introduced in \cref{sec:3}. Note that relaxing the connector angle from the fixed $\pi/2$ to $\theta_e \in [\pi/2, 3\pi/2]$ does not change the underlying topology $\mathcal{G}$ of each configuration $\mathcal{C}$. We exploit the fact that $\mathcal{G}$ is a tree, where the path between any two vertices is unique, and every non-root vertex has exactly one parent\cite{bondy1976graph}. Consequently, a traversal algorithm such as Breadth-First Search (BFS) visits each vertex exactly once, allowing us to assign the connector angle on each edge sequentially and express the full 3D configuration relative to a chosen root module. Since a full traversal touches each vertex and edge a constant number of times, the choice of root affects only the visitation order and not the computational complexity. Therefore, for simplicity we choose the first upper-left module in the representation matrix as the root module. 

We now formalize this. As shown in \cref{fig:design}, the inertial frame $\mathcal{F}_W$ is denoted as the space spanned by three unit basis vectors $\boldsymbol{e}_1 = \left[1, 0, 0 \right]^\top$, $\boldsymbol{e}_2 = \left[0, 1, 0 \right]^\top$, $\boldsymbol{e}_3 = \left[0, 0, 1 \right]^\top$. We place the root module temporarily at the origin $\boldsymbol{p}_1 = [0,0,0]^{\top}$. Its orientation relative to the world frame can be represented as $\boldsymbol{R}_1 = \boldsymbol{R}_x(\alpha_0)\boldsymbol{R}_y(\alpha_1)$, where $\alpha_0,\alpha_1 \in [-\pi/2,\pi/2]$, and $\boldsymbol{R}_{x}(\alpha_i)$ denotes a basic rotation by an angle $\alpha_i$ around the $x$-axis. By applying BFS, all modules' world frame coordinates and orientations can be calculated according to their parent vertices as 
\begin{align*}
    \boldsymbol{R}_i&= 
\begin{dcases}
    \boldsymbol{R}_{i-1}\boldsymbol{R}_{\boldsymbol{x}}(\alpha_{i}),& \text{if } c_v \; \text{or} \; c_h \in \{1, 3\},\\
    \boldsymbol{R}_{i-1}\boldsymbol{R}_{\boldsymbol{y}}(\alpha_{i}),              & \text{otherwise},
\end{dcases}\\
    \boldsymbol{p}_i &= \boldsymbol{p}_{i-1} + l_c\boldsymbol{R}_{i-1}\boldsymbol{t}_i + l_c\boldsymbol{R}_i\boldsymbol{t}_i,
\end{align*}
where $l_c$ is the distance from the COM to a connector, and $\boldsymbol{t}_i$ is a unit vector from the module COM to its connector in its body frame, which depends on $c_v$. With this, the set of $\{\boldsymbol{p}_1, \dots,\boldsymbol{p}_n\}$ and $\{\boldsymbol{R}_1,\dots,\boldsymbol{R}_n\}$ can be computed. Once every module is visited once, we compute the COM position $\boldsymbol{p}_0 = 1/n\sum_{i=1}^n \boldsymbol{p}_i$ and move the structure to the origin, so that each module $i$ position is $\boldsymbol{o}_i = \boldsymbol{p}_i - \boldsymbol{p}_0$. Each module has four rotors, where the relative position of the four rotors is the same in each module body frame; their positions $\boldsymbol{o}_{iq}, q \in \{1,2,3,4\}$ can be computed using $\boldsymbol{p}_i,\boldsymbol{R}_i$ and module arm length $l_{arm}$. For a given angular velocity $\Omega_{iq}$ of each rotor, a thrust of $f_{iq}= c_f \Omega_{iq}^2 $ and a torque of $\tau_{iq}= (-1)^q c_m \Omega_{iq}^2 $ can be generated, where $c_f$ and $c_m$ are rotor constants. 
The overall achievable wrench of the system $\boldsymbol{W} \in \mathbb{R}^6$, comprising the thrust $\boldsymbol{F}$ and the torque $\boldsymbol{\tau}$, is
\begin{equation*}
    \boldsymbol{W} = \left[ \begin{array}{c} \boldsymbol{F} \\ \boldsymbol{\tau}  \end{array} \right] = 
         \left[ \begin{array}{c} \sum_{iq} f_{iq}  \boldsymbol{R}_i \boldsymbol{e}_3 \\ \sum_{iq} f_{iq} \boldsymbol{o}_{iq} \times \boldsymbol{R}_i \boldsymbol{e}_3 + \tau_{iq} \boldsymbol{R}_i \boldsymbol{e}_3 \end{array} \right].
    \label{eq: T and M}
\end{equation*}
For simplicity, we denote by $\boldsymbol{u}_c = \left[\Omega_{11}^2, \dots, \Omega_{n4}^2 \right]^ \top$ the control input. Then the overall wrench $\boldsymbol{W}$ can be expressed as $ \boldsymbol{W} =  \boldsymbol{A} \boldsymbol{u}_c$, with $\boldsymbol{A} =  \left[ \boldsymbol{A}_{11}, \dots , \boldsymbol{A}_{n4} \right]$, where
\begin{equation*}
\boldsymbol{A}_{iq} =  \left[ \begin{array}{c}
c_f\boldsymbol{R}_i \boldsymbol{e}_3  \\
c_f\boldsymbol{o}_{iq} \times \boldsymbol{R}_i \boldsymbol{e}_3  + (-1)^q c_m \boldsymbol{R}_i \boldsymbol{e}_3 \end{array} \right].
\end{equation*}

\subsection{Optimization for a single configuration} \label{sec:opti_one_config}
We adapt the wrench requirement method proposed in \cite{10160555}, for a set of wrenches, if those wrenches are within the achievable polytope of the system, we consider that the wrench requirement is fulfilled. Mathematically, given a set of task wrenches $\{\boldsymbol{b}_k\}_{k=1}^K$, this can be formulated as
\begin{equation*}
\boldsymbol{A}\,\boldsymbol{u}_k=\boldsymbol{b}_k,
\end{equation*}
with element-wise input bounds $0\le \boldsymbol{u}_k \le u_{max}$, where $\boldsymbol{u}_k$ is the input vector used to realize task wrench $\boldsymbol{b}_k$. The total control effort for all tasks can be formulated as $J_{cost} =\sum_{k=1}^K \lambda_k \,\|\boldsymbol{u}_k\|_2^2$. The weights $\lambda_k\!\ge\!0$ reflect task importance, for example, rare tasks may receive smaller weights. Without loss of generality, we set \(\lambda_k = 1\) for all \(k=1,\dots,K\).

To avoid downwash interference, we require a clearance between any two modules $i\neq j$. We propose to use a capsule, which is a cylinder with two hemispherical ends, as a convex hull to enclose the critical collision-free downwash airflow volume for each module: $\boldsymbol{d}_i = \boldsymbol{o}_i - \mu \boldsymbol{R}_i \boldsymbol{e}_3, \; \mu \in [a,b]$, where $\boldsymbol{d}_i$ is a point on the line segment along the $i$-th module's body frame $z$-axis. We use a capsule for its simplicity for collision checking, as the shortest Euclidean distance between two capsules is the minimum distance between their capsule axes, which are two line segments. Although the capsule does not completely capture the fully induced airflow under a quadrotor, by setting the fixed line-segment range $[a,b]$ and the radius $r$ for the hemispherical ends, a wide range of noncollision requirements can be satisfied. For each pair of modules $ i \neq j$, we impose
\[
\|\boldsymbol{d}_i-\boldsymbol{d}_j\|_2^2 \ge 4r^2,
\qquad 1\le i<j\le n,
\]
which yields $n(n-1)/2$ pairwise constraints for an $n$-module configuration. Let $\boldsymbol{\alpha} := [\alpha_0,\dots,\alpha_n]^\top \in \mathbb{R}^{n+1}$, the vector of the root module orientation angles and all relative angles on each connection. Together with the input constraints, we can pose this as a nonlinear optimization problem: 
\begin{alignat}{3} \label{eq:opti}
\min_{\boldsymbol{u}_1,\dots,\boldsymbol{u}_k,\;\boldsymbol{\alpha}} &\quad \sum_{k=1}^K \lambda_k\|\boldsymbol{u}_k\|^2_2 \\ \notag
\text{s.t. } &\quad \|\boldsymbol{d}_i-\boldsymbol{d}_j \|_2^2 \geq 4 r^2, \quad\forall 1\le i<j\le n\\ \notag
&\quad \boldsymbol{A}(\boldsymbol{\alpha}) \boldsymbol{u}_k=\boldsymbol{b}_k, \quad \forall k \in K\\ \notag
&\quad 0 \leq \boldsymbol{u}_k \leq u_{max}, \quad \forall k \in K\\ \notag
&\quad -\pi/2 \leq \boldsymbol{\alpha} \leq \pi/2 \notag
\end{alignat}
where $\boldsymbol{u}_k, \boldsymbol{\alpha}$ are bounded element-wise. We note that, although the objective is quadratic, the equality-constraint matrix $\boldsymbol{A}$ depends on the decision vector $\boldsymbol{\alpha}$, which makes the problem a nonconvex nonlinear program. Global optimization can therefore be computationally expensive. We thus solve it using the gradient-based solver \texttt{Ipopt} within the \texttt{CasADi} toolbox \cite{andersson2019casadi}. \texttt{Ipopt} implements a large-scale interior-point method with a filter line-search and reliably finds locally optimal solutions even for large $n$ \cite{wachter2006implementation}.
\begin{figure*}
    \centering
    \includegraphics[width=1.0\linewidth]{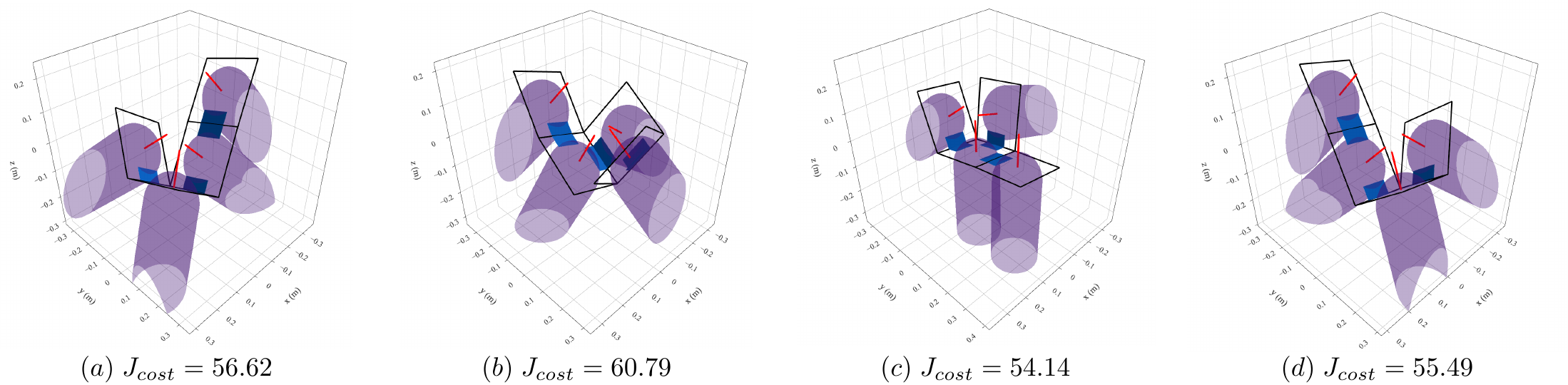}
    \caption{All optimal configurations with control costs $J_{cost}$ using $4$ modules for the wrench set specified in \cref{sec:4-select}.}
    \label{fig:non-iso-opti}
\end{figure*}

\subsection{Selection across configurations}\label{sec:4-select}
Now we provide the final step for our pipeline. Since a target wrench set $\mathcal{W}$ is typically not directly related to the number of modules, we add the module's weight $mg$ scaled by $n$ to each wrench in the $f_z$ direction to compensate for gravity, and we include one more wrench $[0,0,nmg,0,0,0]^{\top}$ to emphasize the assembly hover capability. With the modified wrench target $\mathcal{W}'$, we start at one module and solve the nonlinear programming from \cref{sec:opti_one_config}. For each $n$, we solve all configurations in the given list from \cref{sec:3}, if there is no feasible solution, we increment to $n+1$. When a feasible solution is found, we compare all the costs within this number of modules and select the configuration with the smallest objective value and report the corresponding design. The algorithm is given in \cref{algo:opti}.
\begin{algorithm}
\caption{Selection across configurations}\label{algo:opti}
\begin{algorithmic}[1]
\STATE \textbf{Init:} $n,\{\mathcal{S}_1,\dots,\mathcal{S}_n\}, J_{best}=+\infty, \mathcal{C}_{best}= \{ \emptyset \}$
\FOR{$i=1$ \TO $n$}
    \IF{$J_{best} < +\infty$}
        \STATE \textbf{return} optimal solution found: $\mathcal{C}_{best}$
    \ENDIF
    \FOR{$m=1$ \TO $N_n$}
        \STATE solve \cref{eq:opti} for $\mathcal{C}_m(i)$
        \IF{\text{feasible solution exist}}
        \IF{$J_{cost} < J_{best}$}
            \STATE $J_{best} \gets J_{cost}, \; \mathcal{C}_{best} \gets \mathcal{C}_m(i)$ 
        \ENDIF
        \ENDIF
    \ENDFOR
\ENDFOR
\STATE \textbf{return} no feasible solution found within $n$.
\end{algorithmic}
\end{algorithm}

As an illustration, we solve for a simple wrench set $\mathcal{W}$ of three target wrenches, where all $\boldsymbol{b}_{k,\tau}=\boldsymbol{0}$, with a force set $\{[0.5nmg,0,nmg]^{\top},[0,0.5nmg,nmg]^{\top}, [0,0,nmg]^{\top}\}$ for $4$ modules. We use the parameters in \cref{tab:model-params}, the same as in the \cref{sec:6} real world experiment. We choose the capsule variable $\mu \in [r/2,r/2+10l_{arm}]$, where the line segment is $10$ times the arm length, for a large volume covering the downwash airflow. All configurations with feasible solutions are shown in \cref{fig:non-iso-opti}. As can be seen, the airflow interference constraints are satisfied in all configurations. The cost, which is the sum of equally weighted control input magnitudes to generate all target wrenches, differs across non-isomorphic configurations, which demonstrates the necessity of our two-step approach. The best configuration for $n=12$ is shown in \cref{fig:fig1}, using the target force set $\{[0.1nmg,0,nmg]^{\top},[0,0.1nmg,nmg]^{\top}, [0,0,nmg]^{\top}\}$ with $\boldsymbol{b}_{k,\tau}=\boldsymbol{0}$.

\begin{table}[t]
\centering
\caption{Single module parameters used in this work.}
\label{tab:model-params}
\footnotesize
\setlength{\tabcolsep}{3pt}
\renewcommand{\arraystretch}{0.9}
\begin{threeparttable}
\begin{tabular}{@{}cccccc@{}}
\toprule
$m$ & $c_f$ & $c_m$  & $l_{\text{arm}}$ & $l_{c}$ & $r$\\
\midrule
\SI{0.24}{\kilogram} &
$3.87\times 10^{-7}$ &
$1.06\times 10^{-8}$ &
\SI{0.06}{\metre} &
\SI{0.11}{\metre} &
\SI{0.095}{\metre} \\
\bottomrule
\end{tabular}
\end{threeparttable}
\end{table}

To better illustrate how the overall achievable wrench of the configuration generated by \cref{algo:opti} behaves, we analyze the feasible force polytope \cite{02783649211025998}, which is the set of reachable forces while maintaining zero torque. As can be seen, the target wrenches lie within the boundary of the polytope, which confirms the feasibility of the configuration.

\begin{figure}[t]
    \centering
    \includegraphics[width=1.0\linewidth]{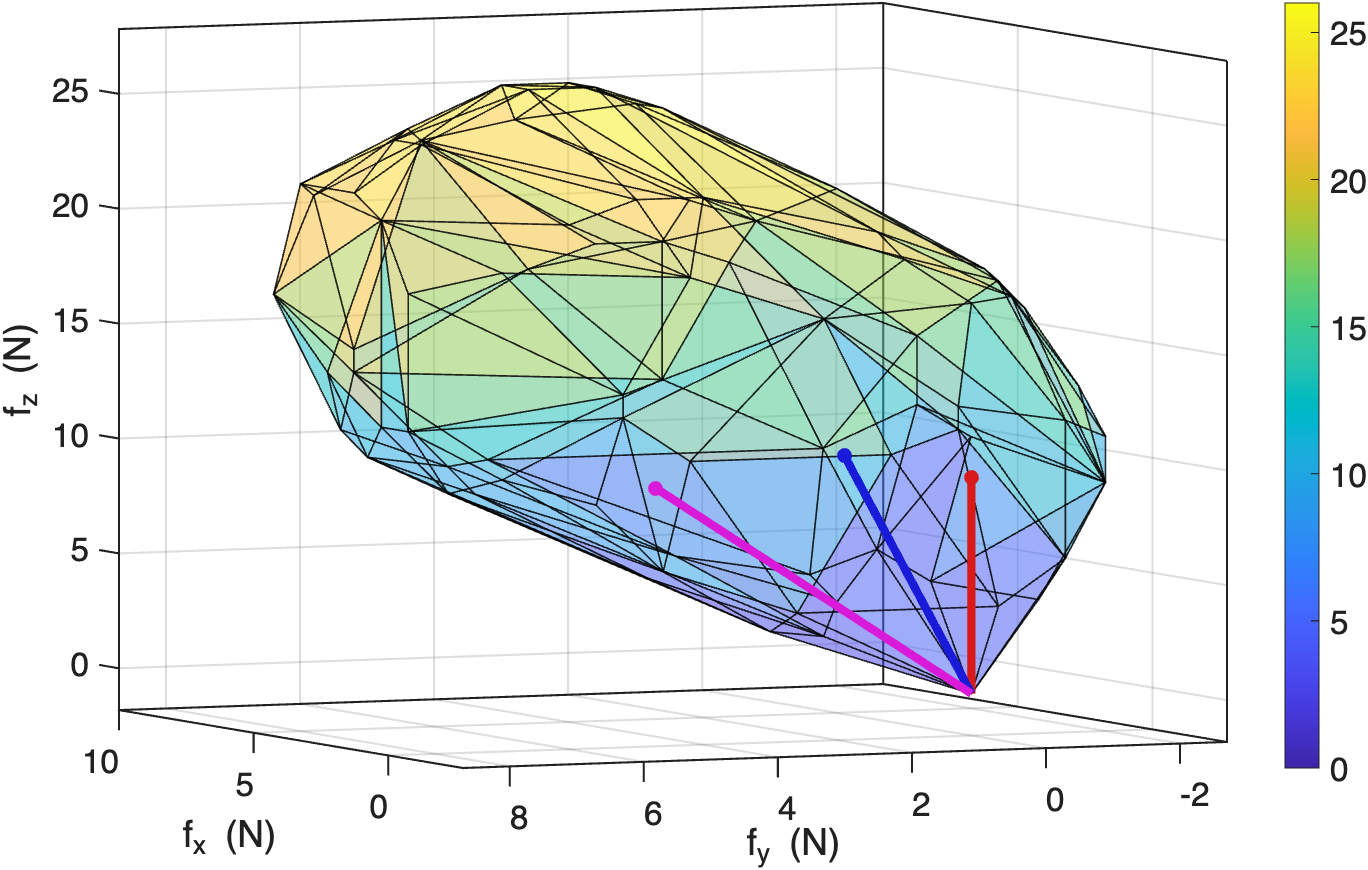}
    \caption{Achievable force polytope for the best configuration in \cref{fig:non-iso-opti} (c); the blue, red, and purple vectors are projections onto the force subspace at zero torque and all lie within the polytope.}
    \label{fig:polytope}
\end{figure}

\begin{figure}
    \centering
    \includegraphics[width=1.0\linewidth]{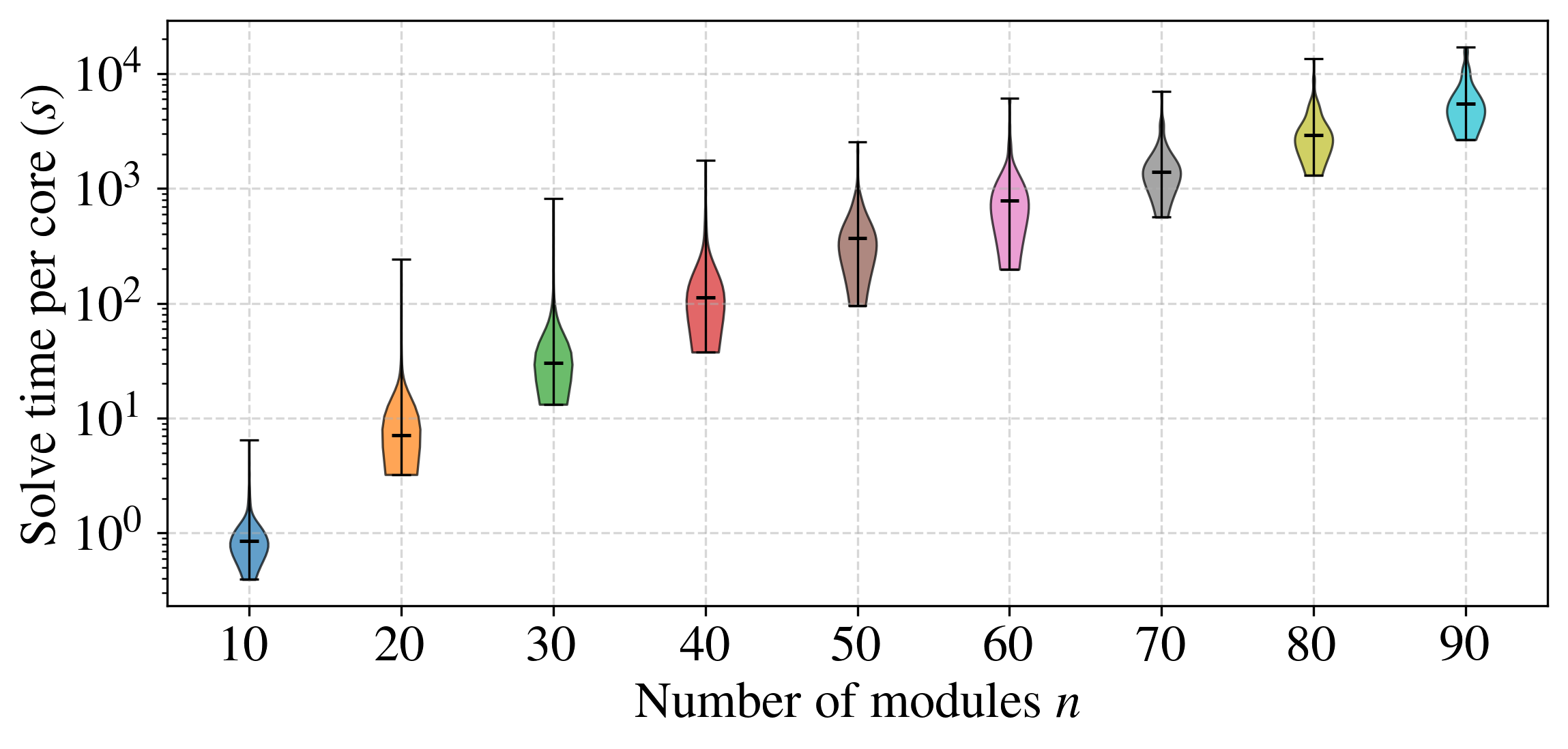}
    \caption{Computation time, averaged per CPU core, with $10$ randomly selected wrenches for $n \in \{10, 20, \dots,90\}$.}
    \label{fig:non-iso-opti-random}
\end{figure}

We also evaluate the computation time for solving \cref{eq:opti}. We randomly select $10$ target wrenches with $f_x,f_y \in [-0.2nmg,0.2nmg]$, $f_z \in [0.9nmg,1.1nmg]$ and target torque components in $[-0.1l_{c}nmg,0.1l_{c}nmg]$. We do not directly apply the wrench set modification in \cref{sec:4-select}, as a target wrench set not related to $n$ might be too large for small $n$ and too trivial for large $n$. The total computation time is given in \cref{fig:non-iso-opti-random}. As an example, for $n=90$ with $10$ target wrenches, the NLP comprises $3691$ decision variables, $60$ equality constraints, and $4005$ inequality constraints. The median single-core solve time is \SI{5460.71}{\second}, demonstrating the scalability of the proposed pipeline. The hardware is the same as in \cref{sec:3}, and with modern hardware the runtime can be further reduced. 

\section{Control} \label{sec:5}
We now derive the control for an optimized configuration with $n$ modules. For its COM position and orientation $( \boldsymbol{p}, \boldsymbol{R}) \in SE (3)$, given a desired trajectory $(\boldsymbol{p}_d, \boldsymbol{R}_d )$, we apply a geometric controller \cite{5717652}, where the desired force and torque can be expressed as
\begin{align*}
\begin{split}
    \boldsymbol{F}_d' &= K_P(\boldsymbol{p}_d-\boldsymbol{p}) + K_D(\dot{\boldsymbol{p}}_d-\dot{\boldsymbol{p}}) + m \ddot{\boldsymbol{p}}_d + mg \boldsymbol{e}_3, \\
    \boldsymbol{\tau}_d  &= - K_R \boldsymbol{e}_R - K_{\omega} \boldsymbol{e}_{\omega} + \boldsymbol{\omega} \times \boldsymbol{J} \boldsymbol{\omega},
    \label{eq: W_d}
\end{split}
\end{align*}
where $\boldsymbol{\omega} \in \mathbb R^3$ is the angular velocity, $K_P, K_D, K_R, K_{\omega}$ are diagonal matrices with positive gains. The error of rotation matrix and the error of angular velocity can be obtained as $\boldsymbol{e}_R = \frac{1}{2} [\boldsymbol{R}_d^\top \boldsymbol{R} - \boldsymbol{R}^\top \boldsymbol{R}_d] ^ \vee $ with $ \boldsymbol{e}_{\omega} =\boldsymbol{\omega} -  \boldsymbol{R}^\top \boldsymbol{R}_d \boldsymbol{\omega}_d$, where $[\cdot]^\vee$ is the inverse of $[\cdot]_\times$, mapping a skew-symmetric matrix to a vector in $\mathbb{R}^3$. The desired thrust in the body frame is given by $\boldsymbol{F}_d = \boldsymbol{R}^{\top} \boldsymbol{F}_d'$. For configurations which are not fully actuated, the desired orientation $\boldsymbol{R}_d$ and desired torque $\boldsymbol{\tau}_d$ can be adjusted based on $\boldsymbol{F}_d'$\cite{10610414}. The desired wrench can be expressed as $\boldsymbol{W}_d =[ \boldsymbol{T}_d^\top, \boldsymbol{\tau}_d^\top]^\top$. The mapping from $\boldsymbol{W}_d$ to rotor inputs can then be formulated as a regularized optimization problem 
\begin{equation*}
    \min_{\boldsymbol{u}_c} \Vert \boldsymbol{A} \boldsymbol{u}_c - \boldsymbol{W}_d \Vert^2_2 + \delta \Vert  \boldsymbol{u}_c \Vert^2_2,
\end{equation*}
where $\delta > 0$ is the regularization constant. The control input has the closed form 
\begin{equation*}
  \boldsymbol{u}_c = \boldsymbol{A}^\top (\boldsymbol{A} \boldsymbol{A}^\top + \delta \boldsymbol{I} )^{-1} \boldsymbol{W}_d,  
  \label{eq: u*}
\end{equation*}
where $\boldsymbol{I} \in \mathbb{R}^{6 \times 6}$ is the identity matrix. Since the motor thrust is limited, we bound the control input to the motor limits. 
\section{Experiments}\label{sec:6}
In this section, we first verify the flight performance of two generated configurations from \cref{fig:non-iso-opti-random} using the real-time physics-based simulation engine \texttt{Bullet}. We then present a real world toy example to further confirm the feasibility of the proposed pipeline.

\subsection{Simulation experiments}
\begin{figure}[t]
    \centering
    \includegraphics[width=1.0\linewidth]{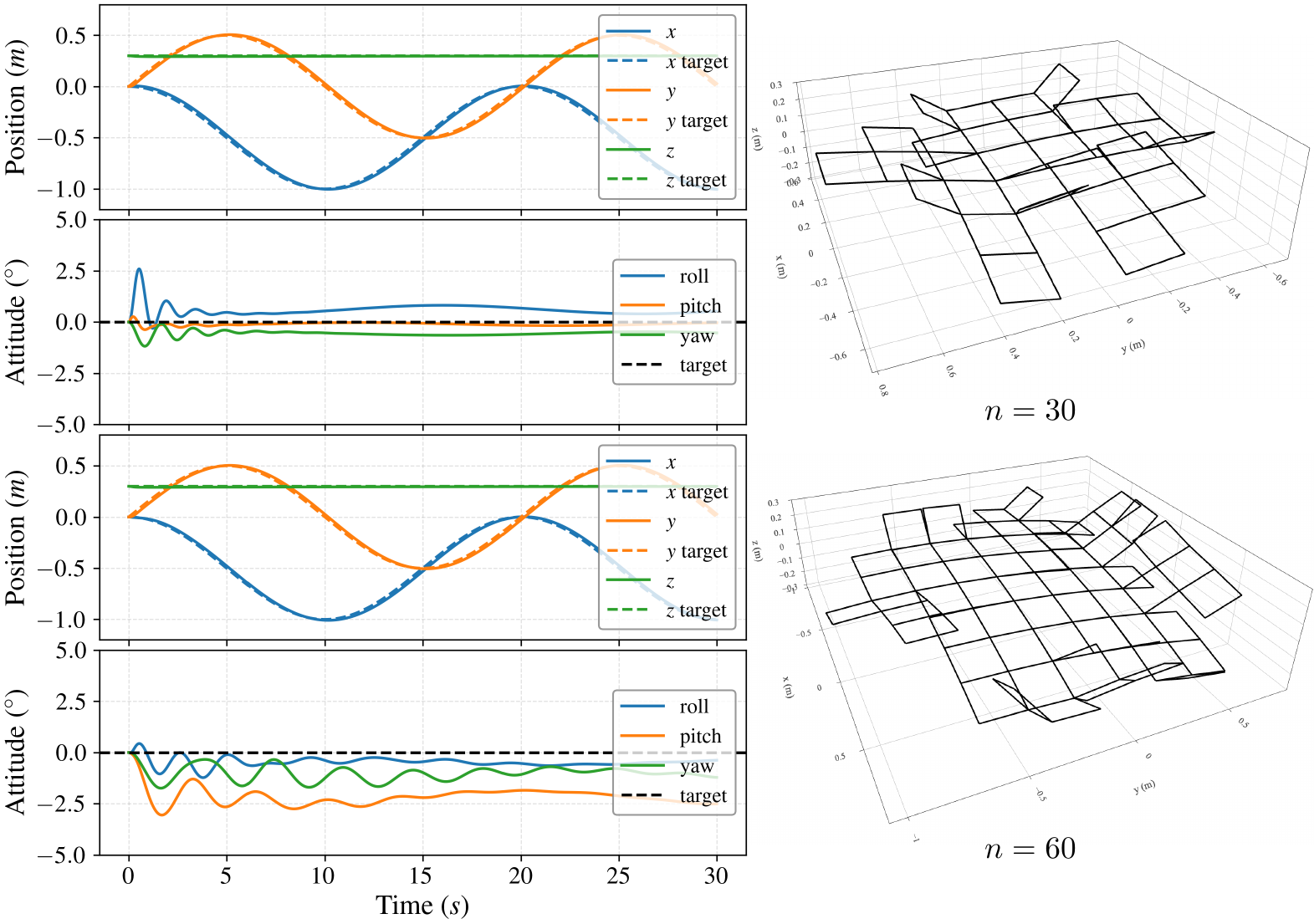}
    \caption{Two optimal configurations from \cref{fig:non-iso-opti-random} with $n=30$ and $n=60$, tracking a circular trajectory with fixed identity orientation.}
    \label{fig:sim-flight}
\end{figure}

Using the controller in \cref{sec:5}, we are able to control a configuration to follow a given trajectory. We choose two of the optimal configurations with $30$ and $60$ modules, obtained in \cref{fig:non-iso-opti-random} using the $10$ randomly generated target wrench set. The flight results are shown in \cref{fig:sim-flight}. For clarity, we plot the 3D illustration with black squares only. Both assemblies track a circular trajectory, with relatively low overall orientation error within $3^{\circ}$. The tracking performance can be further improved by fine-tuning the control parameters in \cref{sec:5}.
\subsection{Real world toy case} \label{sec:6-real}
\begin{figure}[t]
    \centering
    \includegraphics[width=1.0\linewidth]{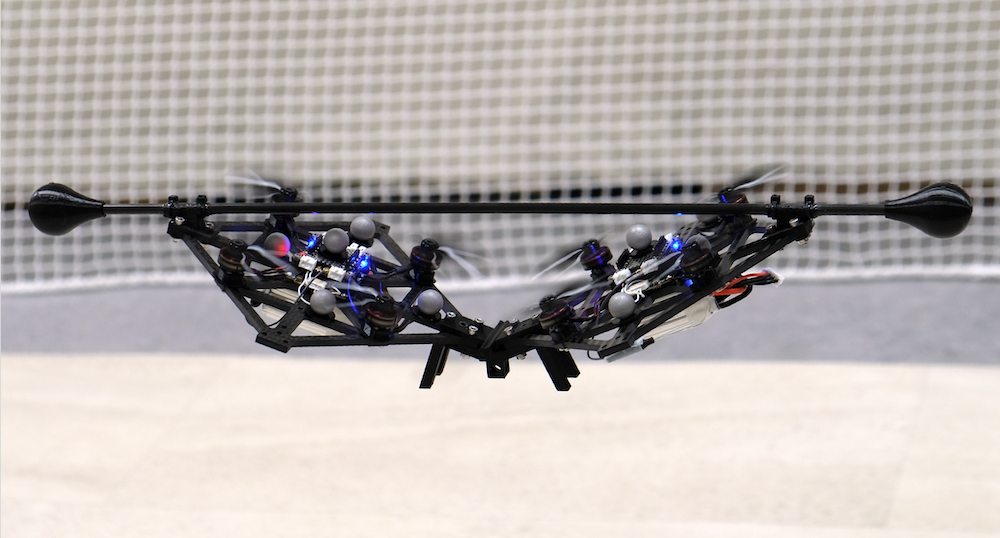}
    \caption{The $2$-module configuration hovering in the air.}
    \label{fig:real-2-modules}
\end{figure}

\begin{figure}
    \centering
    \includegraphics[width=1.0\linewidth]{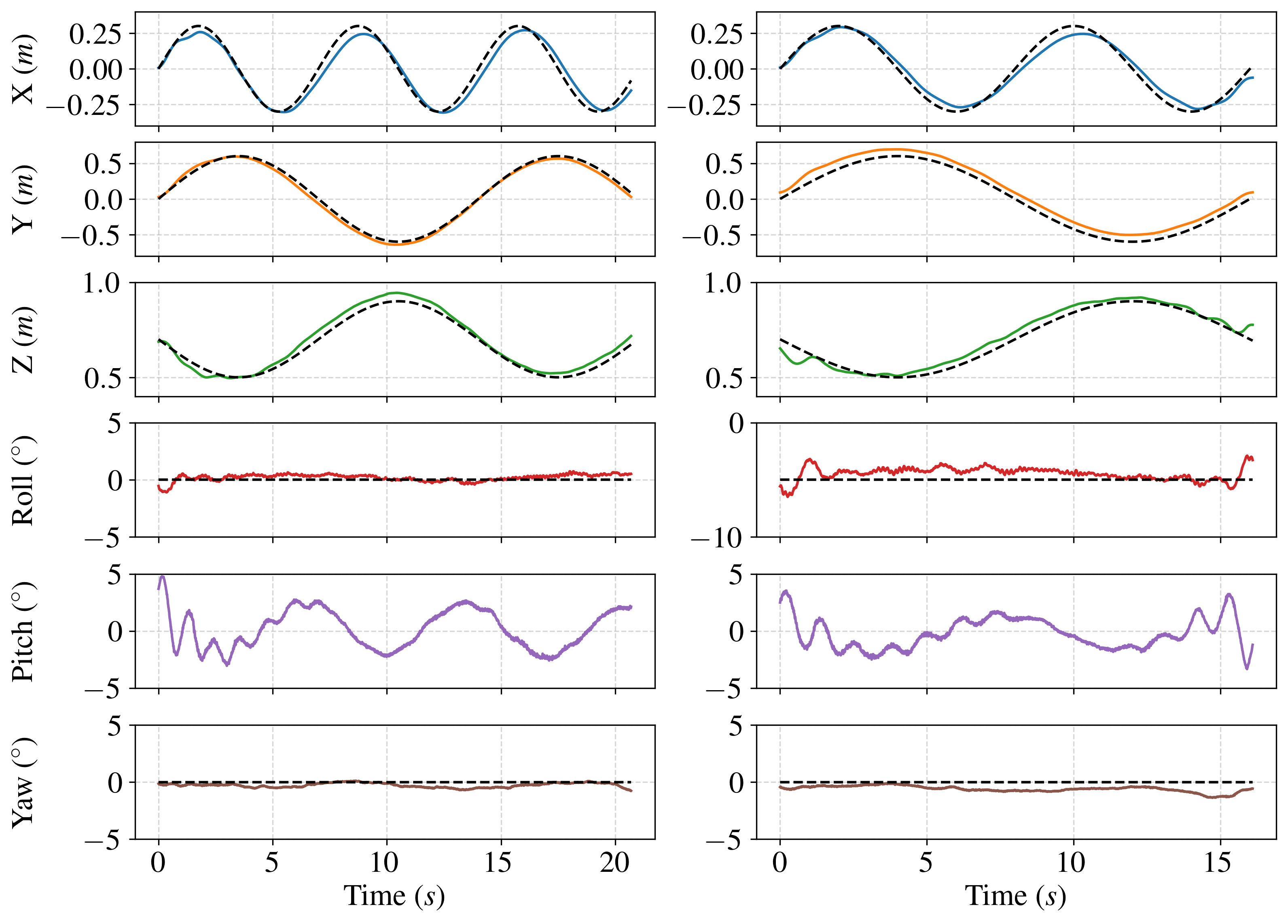}
    \caption{Experiment results of the system in free flight when tracking a 3D figure-$8$ trajectory while maintaining the target roll angle at $0^{\circ}$(left) and $-5^{\circ}$(right). The black dashed lines are the target trajectories.}
    \label{fig:tracking figure 8}
\end{figure}

We also conduct a real world toy example. We define a simple push task along the horizontal $y$-axis, with a target force of \SI{0.36}{\newton} along the $y$-axis, we set $\boldsymbol{b}_{k,\tau}=\boldsymbol{0}$, and the modified force set is $\{[0,0.36,nmg]^{\top},[0,-0.36,nmg]^{\top},[0,0,nmg]^{\top}\}$. By solving \cref{eq:opti} as in \cref{algo:opti}, a feasible solution exists with $n=2$, with optimal angles $\alpha_0=0^{\circ}, \alpha_1=-22.06^{\circ}, \alpha_2=44.12^{\circ}$. We assemble two modules as in \cref{fig:real-2-modules}, as there is only one connection edge, we 3D-print both connectors for simplicity. Note that it is not necessary to have all $6$ controllable DOF to perform aerial physical interaction \cite{6907782}, as in this case, two modules have $5$ controllable DOF and track a $5$-DOF trajectory, while the last DOF is dynamically coupled with the translational dynamics. During physical interaction, the feedback controller is responsible for keeping the system stable. In \cref{fig:tracking figure 8} we show two free-flight trajectory tracking results of the toy-case prototype. These experiments are performed indoors using a VICON motion capture system for localization. Communication between the aerial system and the VICON system is managed using the Crazyswarm framework \cite{7989376}. 
In both flights, the $2$-module structure follows a 3D figure-$8$ trajectory $\boldsymbol{p}_d(t) = [l_0 \sin(2 \pi t/t_c) \cos(2 \pi t/t_c), l_0 \sin(2 \pi t/t_c), h_0 -l_0/3 \sin(2 \pi t/t_c)]^{\top}$ with fixed roll and yaw angles. As the pitch angle is coupled with the translational dynamics, it is therefore not tracked independently. As the overall structure has $5$ controllable DOF, it can maintain the roll angle at a fixed target. The experiments with $0^{\circ}$ and $-5^{\circ}$ are given in \cref{fig:tracking figure 8}. The tracking results indicate that the system maintains its horizontal orientation at $0^{\circ}$, enabling physical interaction along the horizontal $y$-axis as the target wrench set requires.
\section{CONCLUSIONS}\label{sec:7}
In this work, we propose a novel pipeline to design optimal configurations for modular aerial systems. We use homogeneous quadrotor-based modules and assemble them into non-planar structures while ensuring collision-free inter-module downwash constraints, to achieve efficient, physically feasible optimal designs. The proposed approach is scalable and general, and can be applied to other modular robotic systems. In the future, we plan to explore configurations with more modules and construct various assemblies to perform aerial manipulation tasks.
\bibliographystyle{IEEEtran}
\bibliography{references}

@article{annurev023834,
author = "Seo, Jungwon and Paik, Jamie and Yim, Mark",
title = "Modular Reconfigurable Robotics", 
journal= "Annual Review of Control, Robotics, and Autonomous Systems",
year = "2019",
volume = "2",
number = "Volume 2, 2019",
pages = "63-88",
publisher = "Annual Reviews",
issn = "2573-5144",
type = "Journal Article",
}

@INPROCEEDINGS{5509882,
  author={Oung, Raymond and Bourgault, Frédéric and Donovan, Matthew and D'Andrea, Raffaello},
  booktitle={2010 IEEE International Conference on Robotics and Automation}, 
  title={The Distributed Flight Array}, 
  year={2010},
  volume={},
  number={},
  pages={601-607},
  doi={10.1109/ROBOT.2010.5509882}}

@INPROCEEDINGS{8461014,
  author={Saldaña, David and Gabrich, Bruno and Li, Guanrui and Yim, Mark and Kumar, Vijay},
  booktitle={2018 IEEE International Conference on Robotics and Automation (ICRA)}, 
  title={Mod{Q}uad: The Flying Modular Structure that Self-Assembles in Midair}, 
  year={2018},
  volume={},
  number={},
  pages={691-698},
  doi={10.1109/ICRA.2018.8461014}}

@article{0278364919856694,
author = {Markus Ryll and Giuseppe Muscio and Francesco Pierri and Elisabetta Cataldi and Gianluca Antonelli and Fabrizio Caccavale and Davide Bicego and Antonio Franchi},
title ={6{D} interaction control with aerial robots: The flying end-effector paradigm},
journal = {The International Journal of Robotics Research},
volume = {38},
number = {9},
pages = {1045-1062},
year = {2019},
doi = {10.1177/0278364919856694}}

@article{0278364920943654,
author = {Mike Allenspach and Karen Bodie and Maximilian Brunner and Luca Rinsoz and Zachary Taylor and Mina Kamel and Roland Siegwart and Juan Nieto},
title ={Design and optimal control of a tiltrotor micro-aerial vehicle for efficient omnidirectional flight},
journal = {The International Journal of Robotics Research},
volume = {39},
number = {10-11},
pages = {1305-1325},
year = {2020},
doi = {10.1177/0278364920943654},
}

@ARTICLE{5569026,
  author={Michael, Nathan and Mellinger, Daniel and Lindsey, Quentin and Kumar, Vijay},
  journal={IEEE Robotics \& Automation Magazine}, 
  title={The {GRASP} Multiple Micro-{UAV} Testbed}, 
  year={2010},
  volume={17},
  number={3},
  pages={56-65},
  doi={10.1109/MRA.2010.937855}}

@INPROCEEDINGS{7139851,
  author={Nikou, Alexandros and Gavridis, Georgios C. and Kyriakopoulos, Kostas J.},
  booktitle={2015 IEEE International Conference on Robotics and Automation (ICRA)}, 
  title={Mechanical design, modelling and control of a novel aerial manipulator}, 
  year={2015},
  volume={},
  number={},
  pages={4698-4703},
  doi={10.1109/ICRA.2015.7139851}}

@ARTICLE{10214628,
  author={Yu, Pengkang and Su, Yao and Ruan, Lecheng and Tsao, Tsu-Chin},
  journal={IEEE Robotics and Automation Letters}, 
  title={Compensating Aerodynamics of Over-Actuated Multi-Rotor Aerial Platform With Data-Driven Iterative Learning Control}, 
  year={2023},
  volume={8},
  number={10},
  pages={6187-6194},
  doi={10.1109/LRA.2023.3304539}}

@ARTICLE{10804051,
  author={Bauersfeld, Leonard and Muller, Koen and Ziegler, Dominic and Coletti, Filippo and Scaramuzza, Davide},
  journal={IEEE Robotics and Automation Letters}, 
  title={Robotics Meets Fluid Dynamics: A Characterization of the Induced Airflow Below a Quadrotor as a Turbulent Jet}, 
  year={2025},
  volume={10},
  number={2},
  pages={1241-1248},
  doi={10.1109/LRA.2024.3518835}}

@INPROCEEDINGS{10160555,
  author={Xu, Jiawei and Saldaña, David},
  booktitle={2023 IEEE International Conference on Robotics and Automation (ICRA)}, 
  title={Finding Optimal Modular Robots for Aerial Tasks}, 
  year={2023},
  volume={},
  number={},
  pages={11922-11928},
  doi={10.1109/ICRA48891.2023.10160555}}

@INPROCEEDINGS{10801469,
  author={Su, Yao and Jiao, Ziyuan and Zhang, Zeyu and Zhang, Jingwen and Li, Hang and Wang, Meng and Liu, Hangxin},
  booktitle={2024 IEEE/RSJ International Conference on Intelligent Robots and Systems (IROS)}, 
  title={Flight Structure Optimization of Modular Reconfigurable {UAV}s}, 
  year={2024},
  volume={},
  number={},
  pages={4556-4562},
  doi={10.1109/IROS58592.2024.10801469}}

@INPROCEEDINGS{10610414,
  author={Li, Mengguang and Cui, Kai and Koeppl, Heinz},
  booktitle={2024 IEEE International Conference on Robotics and Automation (ICRA)}, 
  title={A Modular Aerial System Based on Homogeneous Quadrotors with Fault-Tolerant Control}, 
  year={2024},
  volume={},
  number={},
  pages={8408-8414},
  doi={10.1109/ICRA57147.2024.10610414}}

@INPROCEEDINGS{6907782,
  author={Yüksel, Burak and Secchi, Cristian and Bülthoff, Heinrich H. and Franchi, Antonio},
  booktitle={2014 IEEE International Conference on Robotics and Automation (ICRA)}, 
  title={Reshaping the physical properties of a quadrotor through IDA-PBC and its application to aerial physical interaction}, 
  year={2014},
  volume={},
  number={},
  pages={6258-6265},
  doi={10.1109/ICRA.2014.6907782}}

@article{doi:10.5772/10489,
author = {Jinguo Liu and Yuechao Wang and Shugen Ma and Yangmin Li},
title ={Enumeration of the Non-Isomorphic Configurations for a Reconfigurable Modular Robot with Square-Cubic-Cell Modules},
journal = {International Journal of Advanced Robotic Systems},
volume = {7},
number = {4},
pages = {31},
year = {2010},
doi = {10.5772/10489},
}

@INPROCEEDINGS{6696972,
  author={Stoy, Kasper and Brandt, David},
  booktitle={2013 IEEE/RSJ International Conference on Intelligent Robots and Systems}, 
  title={Efficient enumeration of modular robot configurations and shapes}, 
  year={2013},
  volume={},
  number={},
  pages={4296-4301},
  doi={10.1109/IROS.2013.6696972}}

@INPROCEEDINGS{9196735,
  author={Gabrich, Bruno and Li, Guanrui and Yim, Mark},
  booktitle={2020 IEEE International Conference on Robotics and Automation (ICRA)}, 
  title={Mod{Q}uad-{D}o{F}: A Novel Yaw Actuation for Modular Quadrotors}, 
  year={2020},
  volume={},
  number={},
  pages={8267-8273},
  doi={10.1109/ICRA40945.2020.9196735}}

@book{bondy1976graph,
  title={Graph theory with applications},
  author={Bondy, John Adrian and Murty, Uppaluri Siva Ramachandra},
  volume={290},
  year={1976},
  publisher={Macmillan London}
}

@article{wachter2006implementation,
  title={On the implementation of an interior-point filter line-search algorithm for large-scale nonlinear programming},
  author={W{\"a}chter, Andreas and Biegler, Lorenz T},
  journal={Mathematical programming},
  volume={106},
  number={1},
  pages={25--57},
  year={2006},
  publisher={Springer}
}

@article{02783649211025998,
author = {Mahmoud Hamandi and Federico Usai and Quentin Sablé and Nicolas Staub and Marco Tognon and Antonio Franchi},
title ={Design of multirotor aerial vehicles: A taxonomy based on input allocation},
journal = {The International Journal of Robotics Research},
volume = {40},
number = {8-9},
pages = {1015-1044},
year = {2021},
doi = {10.1177/02783649211025998},
}

@INPROCEEDINGS{5717652,
  author={Lee, Taeyoung and Leok, Melvin and McClamroch, N. Harris},
  booktitle={49th IEEE Conference on Decision and Control (CDC)}, 
  title={Geometric tracking control of a quadrotor {UAV} on {SE}(3)}, 
  year={2010},
  volume={},
  number={},
  pages={5420-5425},
  doi={10.1109/CDC.2010.5717652}}

@INPROCEEDINGS{7989376,
  author={Preiss, James A. and Honig, Wolfgang and Sukhatme, Gaurav S. and Ayanian, Nora},
  booktitle={2017 IEEE International Conference on Robotics and Automation (ICRA)}, 
  title={Crazyswarm: A large nano-quadcopter swarm}, 
  year={2017},
  volume={},
  number={},
  pages={3299-3304},
  doi={10.1109/ICRA.2017.7989376}}

@INPROCEEDINGS{9636086,
  author={Gabrich, Bruno and Saldaña, David and Yim, Mark},
  booktitle={2021 IEEE/RSJ International Conference on Intelligent Robots and Systems (IROS)}, 
  title={Finding Structure Configurations for Flying Modular Robots}, 
  year={2021},
  volume={},
  number={},
  pages={6970-6976},
  doi={10.1109/IROS51168.2021.9636086}}

@article{andersson2019casadi,
  title={Cas{AD}i: a software framework for nonlinear optimization and optimal control},
  author={Andersson, Joel AE and Gillis, Joris and Horn, Greg and Rawlings, James B and Diehl, Moritz},
  journal={Mathematical Programming Computation},
  volume={11},
  number={1},
  pages={1--36},
  year={2019},
  publisher={Springer}
}

@article{abl6259,
author = {Philipp Foehn  and Elia Kaufmann  and Angel Romero  and Robert Penicka  and Sihao Sun  and Leonard Bauersfeld  and Thomas Laengle  and Giovanni Cioffi  and Yunlong Song  and Antonio Loquercio  and Davide Scaramuzza},
title = {Agilicious: Open-source and open-hardware agile quadrotor for vision-based flight},
journal = {Science Robotics},
volume = {7},
number = {67},
pages = {eabl6259},
year = {2022},
doi = {10.1126/scirobotics.abl6259},
}

@INPROCEEDINGS{10161498,
  author={Cui, Kai and Li, Mengguang and Fabian, Christian and Koeppl, Heinz},
  booktitle={2023 IEEE International Conference on Robotics and Automation (ICRA)}, 
  title={Scalable Task-Driven Robotic Swarm Control via Collision Avoidance and Learning Mean-Field Control}, 
  year={2023},
  volume={},
  number={},
  pages={1192-1199},
  doi={10.1109/ICRA48891.2023.10161498}}
\end{document}